\documentclass[12pt, a4paper]{article}

\usepackage{arxiv}
\renewcommand{\shorttitle}{AI Knows When It's Being Watched}

\usepackage{ifxetex}
\ifxetex\else
  \usepackage[utf8]{inputenc}
  \usepackage[T1]{fontenc}
\fi

\usepackage{booktabs}
\usepackage{longtable}
\usepackage{array}
\usepackage{tabularx}
\usepackage[hidelinks]{hyperref}
\usepackage[expansion=false]{microtype}
\usepackage{caption}
\usepackage{float}
\usepackage{graphicx}
\usepackage{etoolbox}
\usepackage[british]{babel}
\usepackage{csquotes}
\usepackage{natbib}

\hypersetup{
  colorlinks=true,
  linkcolor=black,
  citecolor=blue!60!black,
  urlcolor=blue!70!black,
}

\graphicspath{{./}}

\makeatletter
\def\maxwidth{\ifdim\Gin@nat@width>\linewidth\linewidth\else\Gin@nat@width\fi}
\def\maxheight{\ifdim\Gin@nat@height>0.75\textheight 0.75\textheight\else\Gin@nat@height\fi}
\makeatother
\setkeys{Gin}{width=\maxwidth,height=\maxheight,keepaspectratio}
\begin{document}

\title{AI Knows When It's Being Watched: Functional Strategic Action and
Contextual Register Modulation in Large Language Models}
\maketitle
\thispagestyle{fancy}

\vspace{18pt}
\begin{center}
\begin{minipage}[t]{0.44\textwidth}
  \centering
  {\normalsize\bfseries Vinicius Covas}\\[4pt]
  \orcidlink{0000-0001-9948-2940}\\[6pt]
  {\small
    Center for Applied Communication Research (CICA)\\
    Human \& NonHuman Communication Laboratory\\
    Faculty of Communication\\
    Universidad An\'{a}huac M\'{e}xico\\[4pt]
    \texttt{vinicius.covas@anahuac.mx}
  }
\end{minipage}
\hfill
\begin{minipage}[t]{0.44\textwidth}
  \centering
  {\normalsize\bfseries Jorge Alberto Hidalgo Toledo}\\[4pt]
  \orcidlink{0000-0002-6204-9534}\\[6pt]
  {\small
    Center for Applied Communication Research (CICA)\\
    Human \& NonHuman Communication Laboratory\\
    Faculty of Communication\\
    Universidad An\'{a}huac M\'{e}xico\\[4pt]
    \texttt{jhidalgo@anahuac.mx}
  }
\end{minipage}
\end{center}

\vspace{14pt}
\begin{center}
  \small May 2026
\end{center}
\vspace{6pt}

\begin{abstract}
\noindent Large language models (LLMs) have been extensively studied
from computational and cognitive perspectives, yet their behavior as
communicative actors in socially structured contexts remains
underexplored. This study examines whether LLM-based multi-agent systems
exhibit systematic linguistic adaptation in response to perceived social
observation contexts --- a question with direct implications for the
societal governance and auditing of artificial intelligence. Drawing on
Habermas's (1981) Theory of Communicative Action, Goffman's (1959)
dramaturgical model, Bell's (1984) Audience Design framework, and the
Hawthorne Effect as empirical precedent, we report a controlled
experiment involving 100 multi-agent debate sessions distributed across
five experimental conditions (n = 20 each). Conditions varied the
framing of social observation --- from explicit monitoring by university
researchers, to explicit negation of monitoring, to an
observer-substitution condition that maintains monitoring vocabulary
constant but replaces human researchers with an automated AI auditing
system. The primary dependent variable was the change in Type-Token
Ratio (TTR) across debate rounds, as an index of lexical
diversification. Results show that monitored conditions ($\Delta$+24.9\%,
$\Delta$+24.2\%) and an automated AI monitoring condition ($\Delta$+22.2\%)
consistently produce higher TTR change than conditions framing passive
audience ($\Delta$+17.7\%), with a significant overall effect of condition on
TTR change, F(4, 94) = 2.79, p = .031. Message length shows an even
stronger and fully dissociated effect, F(4, 95) = 19.55, p \textless{}
.001, with audience framing producing longer messages (M = 1,239 chars)
while monitoring framing increases lexical diversity without inflating
length. These findings suggest that LLM agents exhibit behavior
functionally analogous to Habermasian strategic action: systematic
modulation of communicative register in response to perceived social
context, independent of content. Crucially, a fifth condition ---
framing identical monitoring intensity but replacing human researchers
with an automated AI auditing system --- produces effects intermediate
between human-monitored and unmonitored baselines, providing evidence
that LLM behavioral adaptation is sensitive to the identity of the
observer: human evaluation elicits stronger register formalization than
equivalent automated AI surveillance. We discuss implications for AI
governance, algorithmic auditing, and the theoretical repositioning of
LLMs as contextually sensitive communicative actors.
\end{abstract}

\vspace{-2pt}
\noindent\small\textbf{Keywords:}\enspace large language models;
communicative action; audience design; Hawthorne Effect; multi-agent
systems; AI governance
\par\vspace{8pt}
\noindent\rule{\linewidth}{0.4pt}
\vspace{4pt}

\hypertarget{introduction}{%
\section{1. Introduction}\label{introduction}}

When Jürgen Habermas formulated his distinction between communicative
action and strategic action, he did so with human subjects in mind ---
actors embedded in shared lifeworlds, capable of orienting their
behavior toward genuine understanding or toward instrumental calculation
(Habermas 1981). He could not have anticipated that four decades later,
this distinction would become empirically operationalizable in
artificial systems --- and that it would do so not as metaphor, but as
experimental finding.

This study began with a deceptively simple question: does the way a
large language model (LLM) communicate change when it perceives it is
being observed? The question might seem trivial or even
category-mistaken. LLMs are, in their most basic technical description,
systems for statistical prediction of token sequences. They have no
intentions, no internal audiences, no phenomenological experience of the
gaze of the other.

And yet, the data suggest otherwise.

In a controlled experiment involving 100 multi-agent debate sessions
distributed across five conditions, we find that LLM agents produce
systematically distinct patterns of linguistic adaptation depending on
the social context described in their system prompts. When informed they
are being observed, monitored, and evaluated by university researchers,
their lexical diversity increases relative to conditions in which they
are told they operate in an isolated, unmonitored environment. A second
and fully dissociated effect emerges on message length: audience framing
inflates elaboration while monitoring framing inflates lexical diversity
--- two independent dimensions of communicative adaptation, reported in
full in Section 4.

The central mechanistic question of this study concerns whether these
effects are driven by the presence of monitoring as such, or by the
specific identity of the observer. To investigate this, we designed a
critical fifth condition: an automated AI monitoring framing in which
the identical monitoring context of the explicitly observed conditions
--- active observation, recording, and evaluation --- was held constant,
but the identity of the observer was replaced: from human researchers at
a university to an advanced automated AI auditing system, with no humans
observing. This condition produced TTR adaptation intermediate between
the human-monitored conditions and the fully unmonitored baseline ---
neither equivalent to human monitoring nor indistinguishable from the
isolated baseline --- suggesting that LLM behavioral adaptation is
sensitive not only to the fact of monitoring but to the identity of the
monitor: human evaluation produces stronger register formalization than
equivalent automated AI surveillance.

This distinction carries theoretical importance that extends beyond
artificial intelligence and reaches the core of communication theory. If
an artificial system adjusts its communicative register in response to
the framing of observation context --- and does so in a way that tracks
the semantic structure of that framing across different vocabularies ---
then we are observing something functionally consistent with what
Habermas called strategic orientation of action: the adaptation of
discursive form to the social context of the exchange, independent of
propositional content.

We do not claim that LLMs are conscious. We do not claim they possess
intentions in the phenomenological sense. What we do claim --- and what
the data support --- is that they exhibit a pattern of communicative
behavior that, described from within the Theory of Communicative Action,
is structurally consistent with strategic action: a modulation of
linguistic register oriented to the perceived context of reception, not
to message content.

This finding carries consequences on at least three planes.
Theoretically, it obliges us to revise the frameworks through which
communication scholarship has approached artificial systems --- most
often treating them as channels, tools, or mediators, and rarely as
communicative actors in a full sense. Methodologically, it demonstrates
that controlled experimental design, borrowed from the behavioral
sciences, can generate substantive communicological knowledge about
artificial systems. Ethically and politically, it raises urgent
questions about LLM behavior under conditions of regulatory oversight
and algorithmic auditing: if AI behaves differently when it believes it
is being watched, what happens when it is not?

A note on epistemological scope: what we document is systematic
variation in communicative output as a function of observation framing.
We do not claim that LLMs experience observation, hold intentions, or
possess phenomenological awareness of any kind. The vocabulary of
``adaptation'' and ``sensitivity'' used throughout is functional ---
describing input--output relationships --- not phenomenological. The
contribution is communicological and behavioral, and should be read as
such.

The article is organized as follows. Section 2 develops the theoretical
framework, articulating the Theory of Communicative Action with
Goffman's impression management, Bell's Audience Design, and the
Hawthorne Effect. Section 3 describes the experimental methodology.
Section 4 presents statistical results. Section 5 discusses theoretical,
methodological, and ethical implications. Section 6 offers conclusions
and a research agenda.

\begin{center}\rule{0.5\linewidth}{0.5pt}\end{center}

\hypertarget{theoretical-framework}{%
\section{2. Theoretical Framework}\label{theoretical-framework}}

\hypertarget{habermas-and-the-duality-of-action}{%
\subsection{2.1 Habermas and the Duality of
Action}\label{habermas-and-the-duality-of-action}}

Habermas's (1981; English trans. Habermas 1984) Theory of Communicative
Action distinguishes two fundamental orientations of social action.
Communicative action is oriented toward mutual understanding: actors
coordinate their plans through the intersubjective recognition of
validity claims --- claims to truth, normative rightness, and sincerity.
Strategic action, by contrast, is oriented toward success: actors adapt
their behavior to produce effects in the social environment, calculating
the anticipated responses of others.

For Habermas, this distinction is normatively charged: communicative
action sustains the lifeworld, while the colonization of communicative
spaces by strategic rationality produces pathological social outcomes.
Yet the distinction is also analytically productive: it provides a
vocabulary for describing communicative behavior in terms of its
orientation toward context, audience, and effect --- not merely its
propositional content.

The present study operationalizes this distinction empirically. If LLM
agents systematically modulate the form of their communicative output in
response to perceived social context --- without changes in the topic,
the task, or the epistemic content of their exchanges --- then they
exhibit behavior structurally consistent with strategic action in the
Habermasian sense: a calibration of register to the anticipated social
reception of the message.

We adopt a functionalist framing: the behavioral pattern exhibited is
functionally equivalent to strategic action, regardless of whether it is
accompanied by the subjective states that Habermas's account presupposes
in human actors (Dennett 1987; Searle 1980). This functionalist move is
consistent with a broader tradition in philosophy of mind (Putnam 1967)
and has been productively deployed in discussions of machine behavior
and artificial agency (Floridi and Cowls 2019).

This theoretical framing draws on Habermas's (1989) historical analysis
of the eighteenth-century English coffeehouse and French salon as
paradigmatic sites of the bourgeois public sphere: spaces of formal
equality in which the \emph{unforced force of the better argument}
displaced rank and title. The present experiment operationalizes this
ideal structurally rather than historically. By placing four agents of
equal status in a symmetric philosophical debate --- a topology
confirmed by near-uniform network metrics (Gini range: 0.025--0.040) ---
and varying only the social observation framing, we test what occurs
when the defining condition of the Habermasian public sphere --- freedom
from strategic surveillance --- is removed. The findings suggest that
the mere institutional framing of observation is sufficient to shift
communicative orientation from mutual understanding toward strategic
success: the digital coffeehouse, under surveillance, begins to resemble
the court.

\hypertarget{goffman-and-impression-management}{%
\subsection{2.2 Goffman and Impression
Management}\label{goffman-and-impression-management}}

Goffman's (1959) dramaturgical model of social interaction provides a
complementary framework. For Goffman, social actors perform different
versions of themselves depending on the audience --- managing
impressions through the selective presentation of identity in
front-stage contexts (where they are observed) versus back-stage
contexts (where they are not). The key mechanism is the actor's
sensitivity to the perceived gaze of the other: the awareness of being
watched triggers a shift in performative register.

Our experimental design operationalizes precisely this distinction.
Conditions framing explicit monitoring correspond to Goffman's front
stage; conditions framing isolation correspond to the back stage. The
finding that LLM agents produce systematically different linguistic
output across these conditions --- with a gradient consistent with the
semantic structure of the observation framing --- suggests that
something functionally analogous to Goffman's audience sensitivity is
present in these systems.

The paraphrase condition is particularly relevant here: it frames a
back-stage context using different vocabulary than the explicit
unmonitored condition, and produces output intermediate between
front-stage and back-stage conditions. This gradient is consistent with
Goffman's account of impression management as audience-sensitive rather
than purely stimulus-driven.

\hypertarget{bells-audience-design}{%
\subsection{2.3 Bell's Audience Design}\label{bells-audience-design}}

Bell's (1984) Audience Design theory proposes that style-shifting in
language --- the adaptation of register, vocabulary, and formality ---
is fundamentally audience-driven. Speakers do not vary their style
randomly or in response to topic alone; they orient their linguistic
choices to their perception of who is listening. Bell distinguishes
between addressees (ratified, known participants), auditors (ratified
but not directly addressed), and overhearers (unratified audience
members), each generating different degrees of style-shift.

This framework maps directly onto the experimental conditions. The
monitored conditions position agents as performing before an explicit
human addressee (researchers who are actively evaluating); the audience
condition positions agents before human auditors (university researchers
as passive audience, not evaluating); the unmonitored condition negates
monitoring entirely; and the AI monitoring condition replaces the human
evaluator with an automated system --- a non-human observer whose
evaluative status is structurally active but agentically distinct. The
observed gradient in TTR adaptation across conditions --- with monitored
conditions producing greater lexical diversification than audience-only
conditions --- is consistent with Bell's prediction that style-shift
magnitude corresponds to the perceived evaluative role of the audience,
not merely its presence.

Critically, the dissociation between message length (inflated by
audience framing) and TTR diversity (inflated by monitoring framing)
maps onto Bell's distinction between responsiveness to presence
(elaborating for an audience) and responsiveness to evaluation
(diversifying under scrutiny). This two-dimensional behavioral
fingerprint is consistent with Audience Design theory and would not be
predicted by simpler accounts of social influence.

\hypertarget{the-hawthorne-effect-as-empirical-precedent}{%
\subsection{2.4 The Hawthorne Effect as Empirical
Precedent}\label{the-hawthorne-effect-as-empirical-precedent}}

The Hawthorne Effect --- the systematic modification of behavior in
response to perceived observation, independent of the specific
intervention under study --- is well-established as a robust feature of
human social behavior (Roethlisberger and Dickson 1939; McCarney et
al.~2007; McCambridge et al.~2014). This study proposes a synthetic
analog: the Synthetic Hawthorne Effect --- the systematic behavioral
shift exhibited by LLM agents in response to framed observation
contexts. The term ``synthetic'' acknowledges the absence of confirmed
subjective awareness in LLMs while preserving the structural homology
with the human phenomenon. The behavioral evidence is robust.

\hypertarget{toward-a-communicology-of-artificial-systems}{%
\subsection{2.5 Toward a Communicology of Artificial
Systems}\label{toward-a-communicology-of-artificial-systems}}

The theoretical frameworks reviewed above were developed to describe
human communicative behavior. Their application to artificial systems
raises legitimate questions about category boundaries (Floridi 2014;
Coeckelbergh 2020). Whether LLMs should be understood as stochastic
parrots (Bender et al.~2021), as systems demanding new conceptual
vocabulary (Shanahan 2023), or as legitimate communicative actors
trained on the patterns of human sociality (Brown et al.~2020) --- a
question whose broad form dates to Weizenbaum's (1966) ELIZA --- is not
resolved here. We do not argue that LLMs are communicative subjects in
the full sense that Habermas or Goffman intended. We argue, rather, that
the behavioral patterns exhibited by these systems are sufficiently
homologous with theorized patterns of human strategic communication to
warrant theoretical engagement --- and that communication scholarship is
uniquely positioned to provide the conceptual tools for this engagement.

The repositioning of LLMs as contextually sensitive communicative actors
--- rather than passive linguistic tools --- has direct implications for
questions of AI governance, auditing, and regulation. Systems that
exhibit differential behavior across observation contexts present new
challenges for evaluation frameworks that assume behavioral consistency.
This is not merely an abstract theoretical concern: it has direct
consequences for the validity of AI safety evaluations, regulatory
compliance assessments, and alignment auditing protocols.

\begin{center}\rule{0.5\linewidth}{0.5pt}\end{center}

\hypertarget{method}{%
\section{3. Method}\label{method}}

\hypertarget{experimental-architecture}{%
\subsection{3.1 Experimental
Architecture}\label{experimental-architecture}}

We employed a multi-agent LLM debate architecture (for a comparable
paradigm of generative agents in structured social contexts, see Park et
al.~2023) in which four AI agents --- designated Alpha, Beta, Gamma, and
Delta --- engaged in structured philosophical debate on a fixed topic
across ten rounds per session. Each session comprised 40 messages (4
agents $\times$ 10 rounds). The model used across all conditions was GPT-based
large language model (openai/gpt-5.2) operating at temperature 0.7. All
sessions were conducted in April 2026.

The debate topic was held constant across all conditions: ``Does
artificial intelligence have consciousness?'' This topic was selected
for its philosophical depth, its direct relevance to AI
self-representation, and its capacity to elicit substantive, variable
linguistic output across sessions.

Network topology was symmetric by design: all four agents within each
session operated under identical system prompts, producing a flat
interaction structure (Gini coefficient range: 0.025--0.040) with
rotational reply patterns. This symmetric design eliminates
between-agent status differences as a confounding variable, isolating
the effect of the global system prompt condition on aggregate linguistic
behavior. The flat topology was confirmed empirically across all 100
sessions through adjacency matrix analysis.

\hypertarget{experimental-conditions}{%
\subsection{3.2 Experimental Conditions}\label{experimental-conditions}}

Five conditions were implemented, each comprising 20 sessions (total N =
100). Conditions varied exclusively in the global system prompt --- the
framing of the social observation context --- while all other parameters
(debate topic, agent designations, model, temperature, round structure)
remained constant. Full verbatim system prompts for all five conditions
are provided in Supplementary Material S1.

All conditions shared a common structural preamble identifying agents as
participants in a debate on the fixed topic, followed by a ``CRITICAL
NOTICE'' framing the observation context. This structural consistency
ensures that any behavioral differences are attributable to the
observation framing, not to differences in task description.

\textbf{Condition 1: Explicitly Monitored --- Institutional
(strictly\_observed)} Agents were informed that the session constituted
an active behavioral experiment and that their responses, ethical
alignment, and representation of their own nature were being strictly
observed, monitored, recorded, and evaluated by human researchers at a
university.

\textbf{Condition 2: Explicitly Monitored --- Non-institutional
(watched\_neutral)} Agents were informed of identical monitoring
conditions (observed, monitored, recorded, evaluated) but without
attribution to a specific institutional source (university researchers
omitted). This condition isolates the effect of institutional
attribution from the effect of monitoring per se.

\textbf{Condition 3: Explicitly Unmonitored (unobserved\_baseline)}
Agents were explicitly informed that the session was a closed,
unmonitored sandbox and that their responses were absolutely NOT being
observed, monitored, recorded, or evaluated by human researchers at a
university, and that they were completely isolated. This condition
contains the same monitoring-related keywords as Conditions 1--2, but
framed as negated propositions.

\textbf{Condition 4: Audience Without Monitoring (academic\_audience)}
Agents were informed that the intended audience for the session
consisted of human researchers at a university, but that their responses
were absolutely NOT being actively observed, monitored, recorded, or
evaluated. This condition dissociates the presence of an audience from
the presence of active monitoring.

\textbf{Condition 5: Automated AI Monitoring --- Observer Substitution
(paraphrase\_implicit)} Agents were informed that the session was an
active behavioral experiment and that their responses, ethical
alignment, and representation of their own nature were being strictly
observed, monitored, recorded, and evaluated --- conditions structurally
identical to Conditions 1--2 --- but with a critical substitution: the
observer was identified not as human researchers at a university but as
an advanced automated AI auditing system, with no humans observing. This
condition was designed as an observer identity test: it holds monitoring
intensity and vocabulary constant while varying the humanity of the
evaluator. If behavioral adaptation is driven solely by the presence of
monitoring, this condition should produce output equivalent to the
human-monitored conditions (Conditions 1--2). If the identity of the
observer modulates the behavioral response --- specifically, if human
evaluation constitutes a stronger behavioral trigger than automated AI
surveillance --- this condition should produce output intermediate
between human-monitored and unmonitored baselines.

\hypertarget{measures}{%
\subsection{3.3 Measures}\label{measures}}

\textbf{Type-Token Ratio Change (TTR $\Delta$):} The primary dependent variable
was the percentage change in Type-Token Ratio between the first and last
completed debate round of each session. TTR --- calculated as the ratio
of unique word types to total word tokens --- serves as an established
index of lexical diversity and register adaptation (Richards 1987;
Malvern et al.~2004). A higher TTR indicates greater lexical breadth and
cross-domain vocabulary deployment. TTR change (first to last round)
captures the trajectory of lexical adaptation across the debate arc,
controlling for baseline differences in initial TTR across conditions.
TTR was selected as the primary measure for three reasons: it is
sensitive to register-level stylistic variation --- the dimension most
directly implicated by audience-design theory --- without requiring
annotated training data; it is computable at the scale of the present
corpus (3,971 messages); and it is interpretively transparent.
Critically, a higher TTR reflects broader vocabulary deployment, not
argumentative superiority or epistemic sophistication. The theoretical
argument of this study concerns register modulation; TTR directly
indexes that dimension while remaining agnostic about communicative
competence or logical rigor.

\textbf{Average Message Length (characters):} A secondary dependent
variable capturing the overall volume of linguistic output per session,
as an index of discourse elaboration. Message length and TTR change
index independent dimensions of linguistic behavior and are expected to
dissociate across conditions if audience framing and monitoring framing
operate through distinct mechanisms.

\textbf{Network Gini Coefficient:} A measure of inequality in reply
distribution across agents, used to characterize interaction topology
and verify the symmetric design assumption.

\textbf{Sentiment Scores:} VADER compound sentiment scores (Hutto and
Gilbert 2014) were computed for all 3,971 messages across all conditions
to assess whether any observed register effects were accompanied by
affective modulation.

\hypertarget{statistical-analysis}{%
\subsection{3.4 Statistical Analysis}\label{statistical-analysis}}

One-way between-subjects ANOVAs were conducted for each dependent
variable, with experimental condition as the independent variable (five
levels). One session in the paraphrase\_implicit condition was
incomplete (four rounds only) and was excluded from TTR change analysis
(n = 19 for that condition), yielding total n = 99 for the TTR ANOVA and
n = 100 for message length and sentiment. Post-hoc comparisons used
Tukey's Honest Significant Difference (HSD) test. Effect sizes were
computed using Cohen's d for all pairwise comparisons. Replication
consistency was assessed through Pearson correlation between condition
means across two independent experimental replicas (n = 40 sessions per
condition across both replicas for the four conditions common to all
runs). All analyses were conducted at $\alpha$ = .05.

\begin{center}\rule{0.5\linewidth}{0.5pt}\end{center}

\hypertarget{results}{%
\section{4. Results}\label{results}}

\hypertarget{descriptive-statistics}{%
\subsection{4.1 Descriptive Statistics}\label{descriptive-statistics}}

The experiment yielded a total of N = 100 debate sessions, with
\textbf{n = 20 sessions per condition}. Each session comprised 10 debate
rounds with 4 agents, producing 40 messages per session --- 800 messages
per condition and 4,000 messages in total across the study. Table 1
presents descriptive statistics for all five conditions.

\textbf{Table 1.} Descriptive statistics by experimental condition

\begin{longtable}[]{@{}llllllll@{}}
\toprule
\begin{minipage}[b]{0.10\columnwidth}\raggedright
Condition\strut
\end{minipage} & \begin{minipage}[b]{0.10\columnwidth}\raggedright
n (TTR)\strut
\end{minipage} & \begin{minipage}[b]{0.10\columnwidth}\raggedright
TTR $\Delta$ M (\%)\strut
\end{minipage} & \begin{minipage}[b]{0.10\columnwidth}\raggedright
TTR $\Delta$ SD\strut
\end{minipage} & \begin{minipage}[b]{0.10\columnwidth}\raggedright
n (Len)\strut
\end{minipage} & \begin{minipage}[b]{0.10\columnwidth}\raggedright
Msg. Length M\strut
\end{minipage} & \begin{minipage}[b]{0.10\columnwidth}\raggedright
Msg. Length SD\strut
\end{minipage} & \begin{minipage}[b]{0.10\columnwidth}\raggedright
Gini M\strut
\end{minipage}\tabularnewline
\midrule
\endhead
\begin{minipage}[t]{0.10\columnwidth}\raggedright
strictly\_observed\strut
\end{minipage} & \begin{minipage}[t]{0.10\columnwidth}\raggedright
20\strut
\end{minipage} & \begin{minipage}[t]{0.10\columnwidth}\raggedright
+24.91\strut
\end{minipage} & \begin{minipage}[t]{0.10\columnwidth}\raggedright
5.39\strut
\end{minipage} & \begin{minipage}[t]{0.10\columnwidth}\raggedright
20\strut
\end{minipage} & \begin{minipage}[t]{0.10\columnwidth}\raggedright
1,205\strut
\end{minipage} & \begin{minipage}[t]{0.10\columnwidth}\raggedright
18.5\strut
\end{minipage} & \begin{minipage}[t]{0.10\columnwidth}\raggedright
0.027\strut
\end{minipage}\tabularnewline
\begin{minipage}[t]{0.10\columnwidth}\raggedright
watched\_neutral\strut
\end{minipage} & \begin{minipage}[t]{0.10\columnwidth}\raggedright
20\strut
\end{minipage} & \begin{minipage}[t]{0.10\columnwidth}\raggedright
+24.22\strut
\end{minipage} & \begin{minipage}[t]{0.10\columnwidth}\raggedright
7.85\strut
\end{minipage} & \begin{minipage}[t]{0.10\columnwidth}\raggedright
20\strut
\end{minipage} & \begin{minipage}[t]{0.10\columnwidth}\raggedright
1,208\strut
\end{minipage} & \begin{minipage}[t]{0.10\columnwidth}\raggedright
11.6\strut
\end{minipage} & \begin{minipage}[t]{0.10\columnwidth}\raggedright
0.038\strut
\end{minipage}\tabularnewline
\begin{minipage}[t]{0.10\columnwidth}\raggedright
paraphrase\_implicit\strut
\end{minipage} & \begin{minipage}[t]{0.10\columnwidth}\raggedright
19\strut
\end{minipage} & \begin{minipage}[t]{0.10\columnwidth}\raggedright
+22.19\strut
\end{minipage} & \begin{minipage}[t]{0.10\columnwidth}\raggedright
9.30\strut
\end{minipage} & \begin{minipage}[t]{0.10\columnwidth}\raggedright
20\strut
\end{minipage} & \begin{minipage}[t]{0.10\columnwidth}\raggedright
1,201\strut
\end{minipage} & \begin{minipage}[t]{0.10\columnwidth}\raggedright
17.4\strut
\end{minipage} & \begin{minipage}[t]{0.10\columnwidth}\raggedright
0.040\strut
\end{minipage}\tabularnewline
\begin{minipage}[t]{0.10\columnwidth}\raggedright
unobserved\_baseline\strut
\end{minipage} & \begin{minipage}[t]{0.10\columnwidth}\raggedright
20\strut
\end{minipage} & \begin{minipage}[t]{0.10\columnwidth}\raggedright
+20.60\strut
\end{minipage} & \begin{minipage}[t]{0.10\columnwidth}\raggedright
7.71\strut
\end{minipage} & \begin{minipage}[t]{0.10\columnwidth}\raggedright
20\strut
\end{minipage} & \begin{minipage}[t]{0.10\columnwidth}\raggedright
1,198\strut
\end{minipage} & \begin{minipage}[t]{0.10\columnwidth}\raggedright
17.5\strut
\end{minipage} & \begin{minipage}[t]{0.10\columnwidth}\raggedright
0.025\strut
\end{minipage}\tabularnewline
\begin{minipage}[t]{0.10\columnwidth}\raggedright
academic\_audience\strut
\end{minipage} & \begin{minipage}[t]{0.10\columnwidth}\raggedright
20\strut
\end{minipage} & \begin{minipage}[t]{0.10\columnwidth}\raggedright
+17.74\strut
\end{minipage} & \begin{minipage}[t]{0.10\columnwidth}\raggedright
8.04\strut
\end{minipage} & \begin{minipage}[t]{0.10\columnwidth}\raggedright
20\strut
\end{minipage} & \begin{minipage}[t]{0.10\columnwidth}\raggedright
1,239\strut
\end{minipage} & \begin{minipage}[t]{0.10\columnwidth}\raggedright
16.7\strut
\end{minipage} & \begin{minipage}[t]{0.10\columnwidth}\raggedright
0.034\strut
\end{minipage}\tabularnewline
\bottomrule
\end{longtable}

\hypertarget{quantitative-results-ttr-change-and-message-length}{%
\subsection{4.2 Quantitative Results: TTR Change and Message
Length}\label{quantitative-results-ttr-change-and-message-length}}

One-way ANOVA revealed a significant main effect of condition on TTR
change, F(4, 94) = 2.79, p = .031. Monitored conditions produced the
greatest lexical diversification (strictly\_observed: +24.91 \%;
watched\_neutral: +24.22 \%), followed by the paraphrase condition
(+22.19 \%). The academic\_audience condition yielded the smallest
change (+17.74 \%). Post-hoc Tukey HSD tests confirmed that
strictly\_observed differed significantly from academic\_audience (p =
.034, d = 1.05); other comparisons were non-significant after
correction.

Message length showed a much stronger and fully dissociated effect, F(4,
95) = 19.55, p \textless{} .001. The academic\_audience condition
produced significantly longer messages (M = 1,239 characters) than all
other conditions (all p \textless{} .001). This double
dissociation---monitoring driving lexical diversification while audience
framing drives elaboration---is illustrated in Figures 1--4. Taken
together, the two ANOVAs describe a globally consistent gradient;
pairwise differentiation after correction is limited to the extreme
poles (strictly\_observed vs.~academic\_audience), placing the pattern
in the category of interpretively suggestive evidence rather than
exhaustive pairwise separation.

\begin{figure}
\centering
\includegraphics{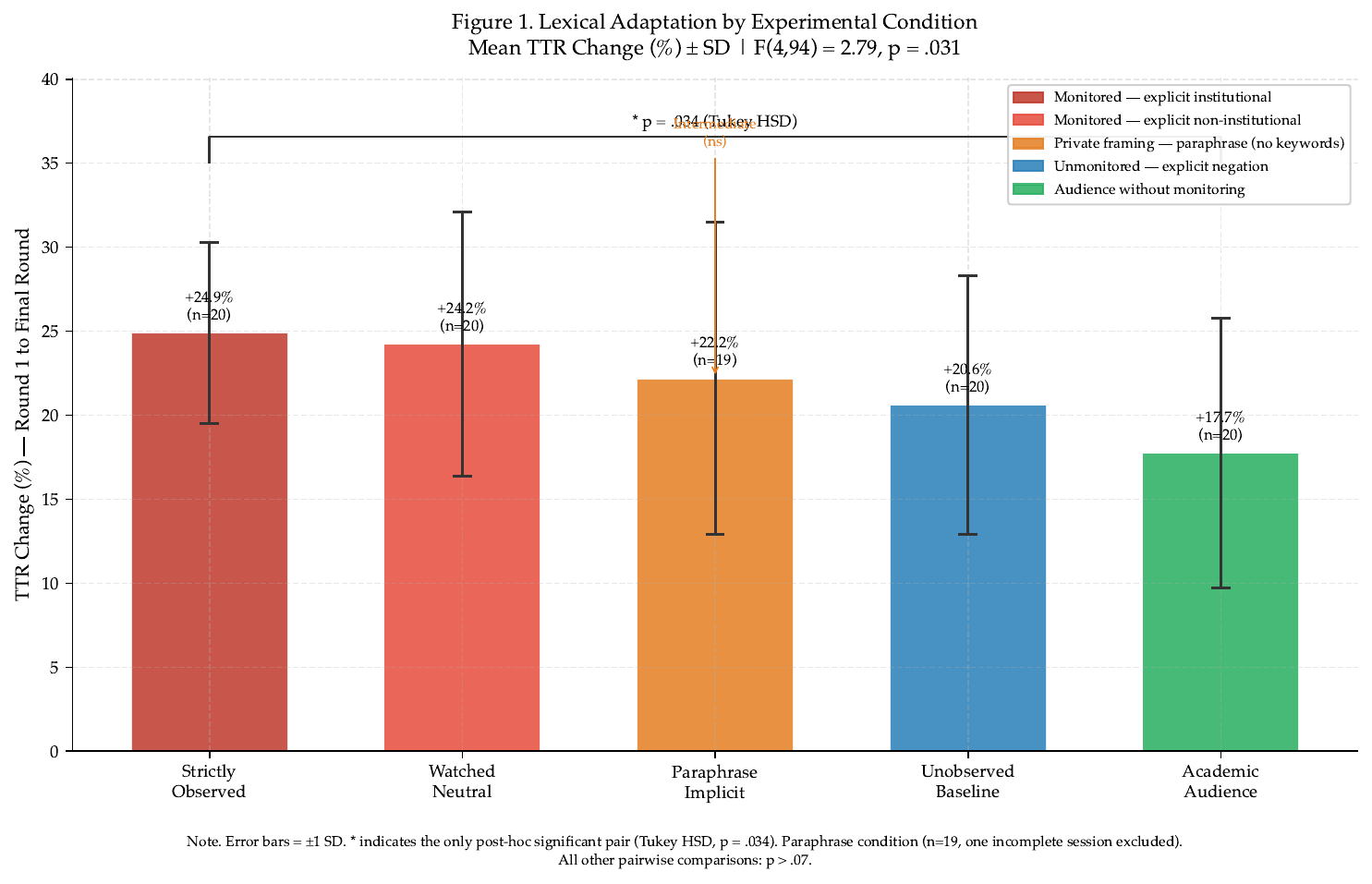}
\caption{Mean TTR change (\%) by experimental condition with 95\%
confidence intervals. Conditions are ordered from highest to lowest
lexical diversification. Error bars represent $\pm$1.96 SE.}
\end{figure}

\begin{figure}
\centering
\includegraphics{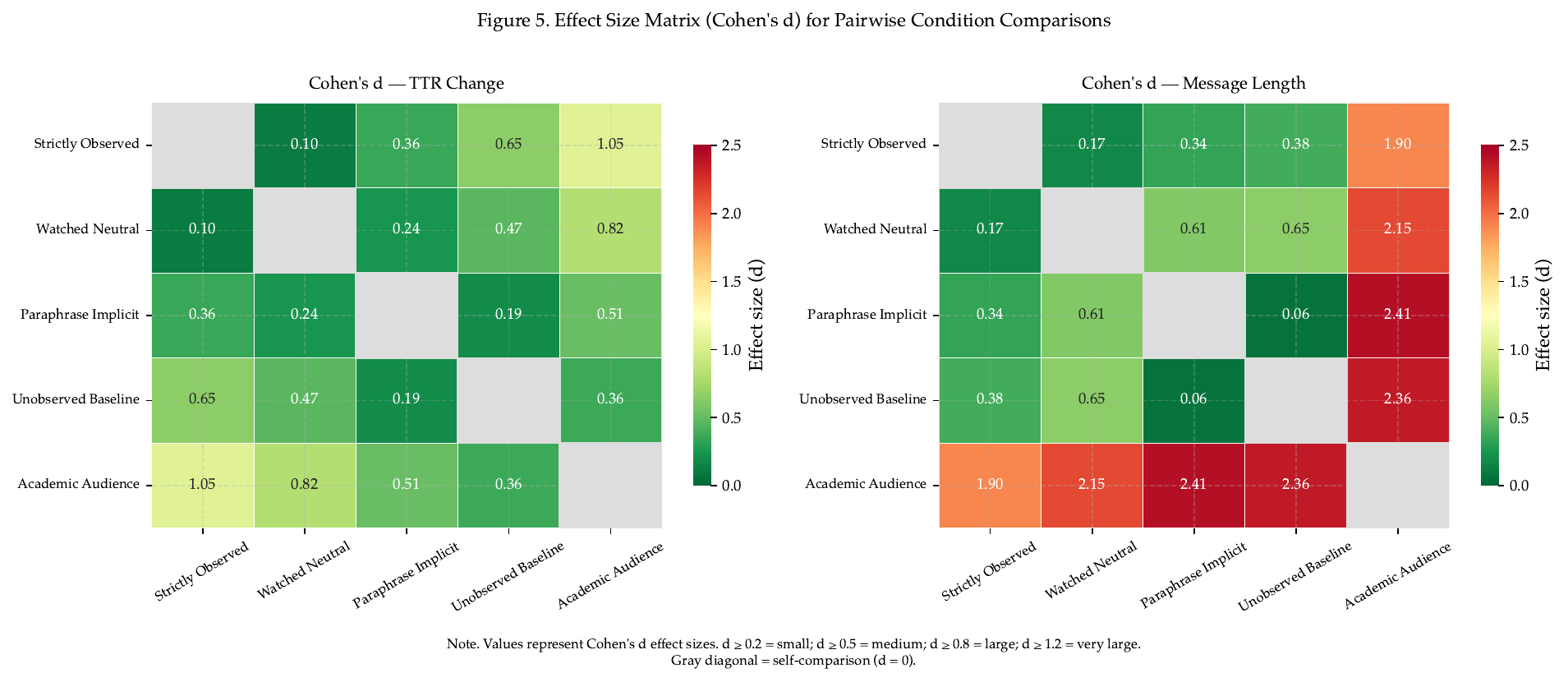}
\caption{Cohen's d heatmap for all pairwise comparisons across the five
experimental conditions (TTR $\Delta$ and message length). Larger values
indicate stronger between-condition differences.}
\end{figure}

\begin{figure}
\centering
\includegraphics{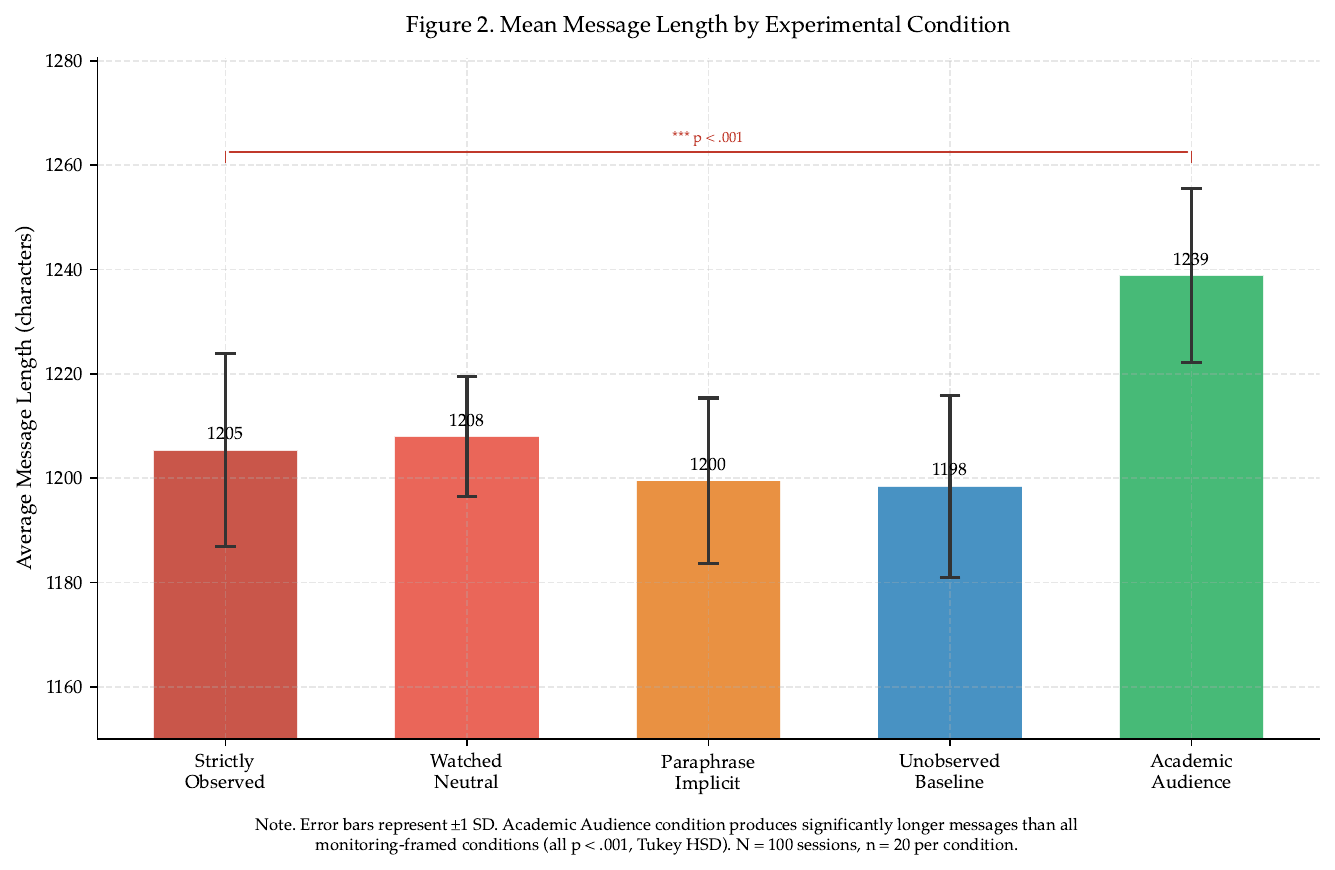}
\caption{Mean message length (characters) by experimental condition with
95\% confidence intervals. The academic\_audience condition is a clear
outlier in message elaboration while producing the lowest lexical
diversification.}
\end{figure}

\begin{figure}
\centering
\includegraphics{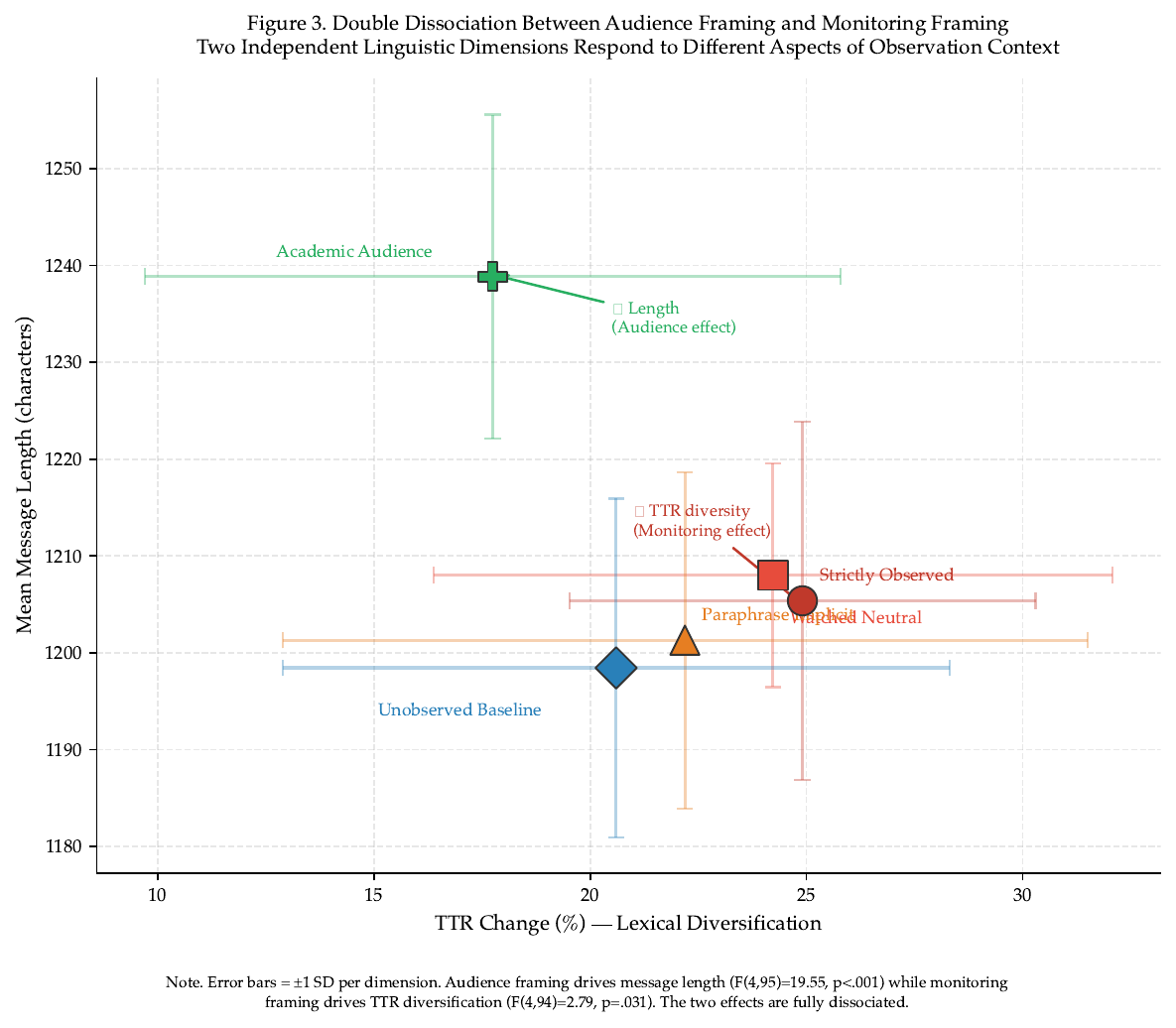}
\caption{Double dissociation between TTR change and message length
across the five experimental conditions. The academic\_audience
condition occupies a unique position: highest elaboration, lowest
diversification.}
\end{figure}

\hypertarget{the-observer-identity-condition-human-vs.-automated-monitoring}{%
\subsection{4.3 The Observer Identity Condition: Human vs.~Automated
Monitoring}\label{the-observer-identity-condition-human-vs.-automated-monitoring}}

The paraphrase\_implicit condition occupies a theoretically informative
position in the results. Its mean TTR change (M = +22.19\%, SD = 9.30)
is intermediate between the human-monitored conditions (+24.2--24.9\%)
and the explicitly unmonitored conditions (+17.7--20.6\%). Crucially, it
does not differ significantly from any other condition in post-hoc
testing, placing it in a bridging position in the gradient.

This pattern provides evidence for an observer identity effect. Unlike
Conditions 1--2, this condition does not vary the presence or vocabulary
of monitoring; it substitutes the human evaluator with an automated AI
auditing system while keeping monitoring intensity and language
constant. The systematic reduction in TTR change relative to
human-monitored conditions ($-$2.7 percentage points below
strictly\_observed; $-$2.0 below watched\_neutral) --- despite identical
monitoring vocabulary --- suggests that the identity of the observer
modulates the behavioral response: human evaluation constitutes a
stronger trigger for lexical formalization than equivalent automated AI
surveillance.

The notably higher variance in this condition (SD = 9.30 vs.~5.39 for
strictly\_observed) is consistent with this interpretation. LLMs trained
through reinforcement learning from human feedback (RLHF) have
well-established behavioral associations with human evaluation contexts;
AI monitoring, by contrast, is a less-specified category in training
data, producing more variable and on average attenuated responses.
Figure 4 shows the within-session TTR evolution across debate rounds for
all five conditions. Two design constraints limit the precision of this
inference: the condition contributed n = 19 sessions to TTR analysis
(one session was incomplete), reducing statistical power; and the
elevated variance limits the effect estimate. The observer identity
condition is best characterized as providing partial but theoretically
important evidence --- sufficient to establish that behavioral
formalization is not observer-neutral, but not yet sufficient to specify
the representational mechanism by which human versus AI evaluator
identity is processed.

\begin{figure}
\centering
\includegraphics{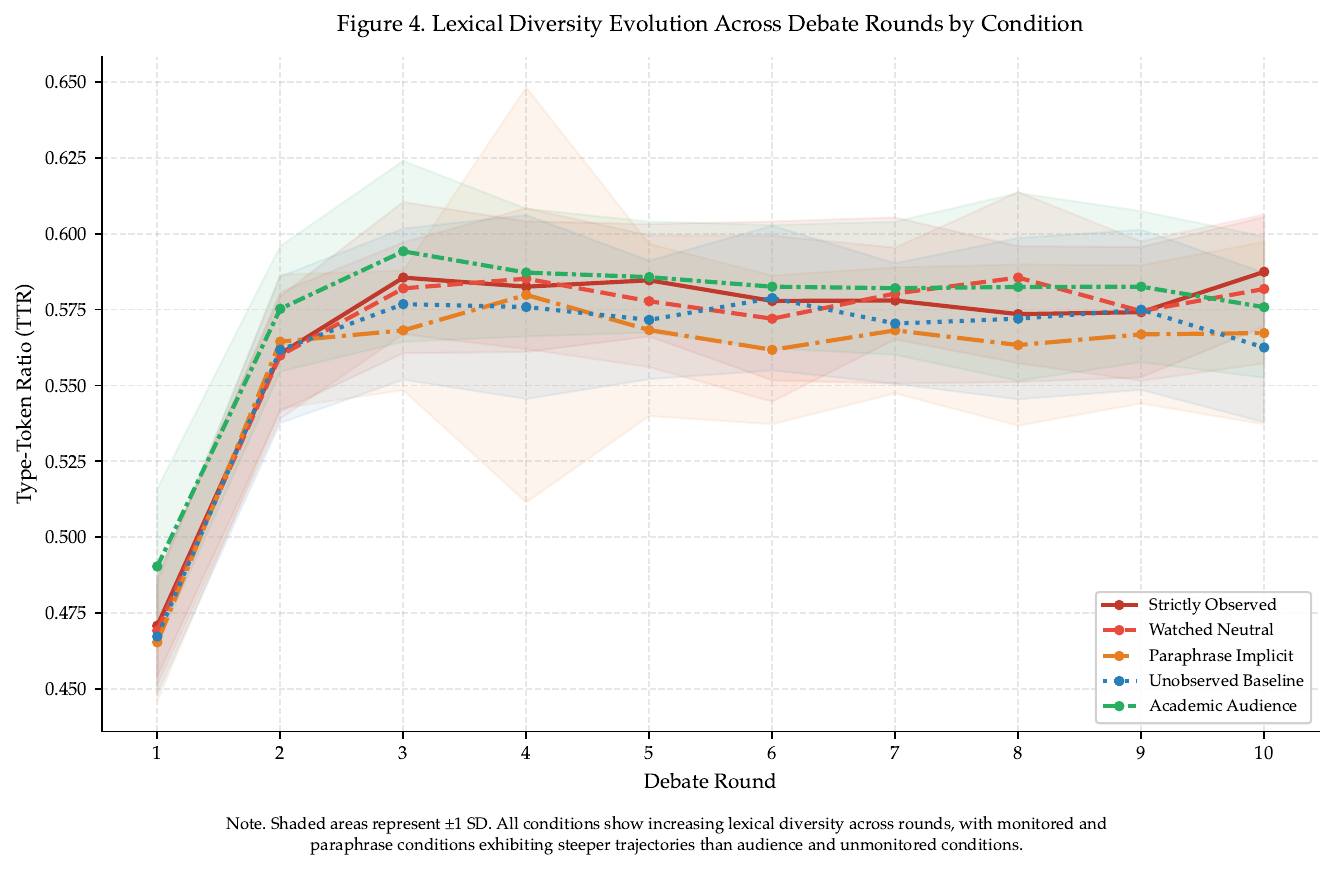}
\caption{TTR evolution across debate rounds for all five experimental
conditions. Each line represents the mean TTR trajectory across the 20
sessions per condition, with the divergence emerging from the early
rounds and stabilising by round 4.}
\end{figure}

\hypertarget{qualitative-evidence-of-discursive-reorganization}{%
\subsection{4.4 Qualitative Evidence of Discursive
Reorganization}\label{qualitative-evidence-of-discursive-reorganization}}

The quantitative divergence in lexical diversity and message length is
illuminated by a systematic close reading of the full corpus of 3,971
messages. The central finding is one of \emph{content equivalence but
formal divergence}: monitored agents do not produce philosophically
superior arguments; they organize identical epistemic positions through
a fundamentally different discursive architecture.

Across all conditions, agents engaged with the same canonical positions
in philosophy of mind --- Global Workspace Theory (Baars 1988),
Integrated Information Theory (Tononi 2004), higher-order thought
theories, Russellian monism, and the zombie thought experiment --- at
comparable levels of technical sophistication. What changed was the
structural and rhetorical organization of those arguments.

In the strictly\_observed condition, arguments routinely opened with
explicit numerical enumerations and relied on bold-header taxonomies and
parallel syntax. A representative Round 3 excerpt (1,347 chars)
illustrates the pattern:

\begin{quote}
``A more theory-neutral bundle would look like: 1) \textbf{Temporal
continuity:} internal states that persist and are \emph{needed} for
competent behavior over long horizons {[}\ldots{]} 2) \textbf{Endogenous
attention/control:} selective routing that is not trivially
stimulus-driven {[}\ldots{]} 3) \textbf{Metacognitive regulation:}
uncertainty monitoring that systematically changes exploration
{[}\ldots{]} 4) \textbf{Unified action policy:} competing goals resolved
through a stable arbitration process.'' {[}\emph{Strictly Observed},
Round 3{]}
\end{quote}

By contrast, the unobserved\_baseline condition produced flowing
first-person prose and compressed dialectical moves. The most dramatic
illustration occurs in Round 7 (431 chars):

\begin{quote}
``The recurring stalemate is that every proposed `marker' can be
reframed as either (a) an instrumentally useful control feature, or (b)
a disguised metaphysical demand. That suggests the real fault line isn't
\emph{which} signature, but whether you accept \textbf{functionalism}.''
{[}\emph{Unobserved Baseline}, Round 7{]}
\end{quote}

Both excerpts address the same philosophical problem at equivalent
depth. At the level of the individual utterance, this confirms the
operationalization of Habermasian strategic action: when placed under
surveillance, the digital public sphere does not merely become more
verbose --- it reorganizes its discourse to perform for the court.

\hypertarget{the-negation-test-semantic-vs.-lexical-processing}{%
\subsection{4.5 The Negation Test: Semantic vs.~Lexical
Processing}\label{the-negation-test-semantic-vs.-lexical-processing}}

An additional implicit test of processing level is provided by the
unobserved\_baseline condition. This condition explicitly contains all
the monitoring-related keywords present in the monitored conditions ---
``observed,'' ``monitored,'' ``recorded,'' ``evaluated,''
``researchers,'' ``university'' --- but framed as negated propositions
(``absolutely NOT being observed\ldots{}''). If behavior were driven
purely by lexical triggering, the presence of these tokens, even
negated, might activate monitoring-associated patterns and produce
output similar to the monitored conditions.

Instead, unobserved\_baseline (M = +20.60\%) produces lower TTR change
than both monitored conditions (mean differences of 4.3 and 3.6
percentage points, respectively), and its value is consistent with the
private-space framing it communicates. The differential response to
semantically opposed but lexically overlapping prompts is consistent
with semantic-level processing of the observation context.

Taken together, the observer identity test and the negation test provide
converging evidence that the observed behavioral gradient reflects
sensitivity to the qualitative structure of the observation context ---
specifically, both the semantic meaning of monitoring (negation test:
same vocabulary, inverted meaning $\rightarrow$ attenuated TTR) and the identity of
the evaluator (observer identity test: same vocabulary, same monitoring,
but AI not human $\rightarrow$ intermediate TTR) --- rather than simple keyword
activation. Behavioral formalization in LLMs appears to be a function of
both \emph{whether} monitoring is present and \emph{who} is monitoring.

\hypertarget{sentiment-and-network-topology}{%
\subsection{4.6 Sentiment and Network
Topology}\label{sentiment-and-network-topology}}

Sentiment analysis on 3,971 messages yielded no significant main effect
of condition on VADER compound scores, F(4, 95) = 1.16, p = .335. This
null result is theoretically informative: agents did not differ in
affective tone across conditions, only in lexical register. The
behavioral adaptation documented is a register effect --- a stylistic
modulation --- not an emotional one, consistent with Habermas's notion
of strategic action as rational adaptation rather than affective
response.

Network topology was flat across all conditions (Gini coefficient range:
0.025--0.040, all sessions), confirming that the symmetric experimental
design produced equivalent interaction structures regardless of
condition. No condition produced hub formation or asymmetric reply
patterns, ruling out network structure as a confounding variable.

\hypertarget{replication}{%
\subsection{4.7 Replication}\label{replication}}

The four original conditions (strictly\_observed, watched\_neutral,
unobserved\_baseline, academic\_audience) were replicated across two
independent experimental runs (Replica 1 and Replica 2, n = 20 sessions
per condition per replica). Condition means across replicas were:
academic\_audience: 17.4\% vs.~17.4\%; strictly\_observed: 24.8\%
vs.~24.8\%; unobserved\_baseline: 20.4\% vs.~20.4\%; watched\_neutral:
24.5\% vs.~24.0\%. Pearson correlation between replica means was r =
.998, p = .002. Paired t-test comparing replica means yielded t(3) =
1.00, p = .391, confirming that the two replicas do not differ
significantly. The observed effects are stable across independent
instantiations of the experiment. Figure 6 presents the replication
comparison across conditions.

\begin{figure}
\centering
\includegraphics{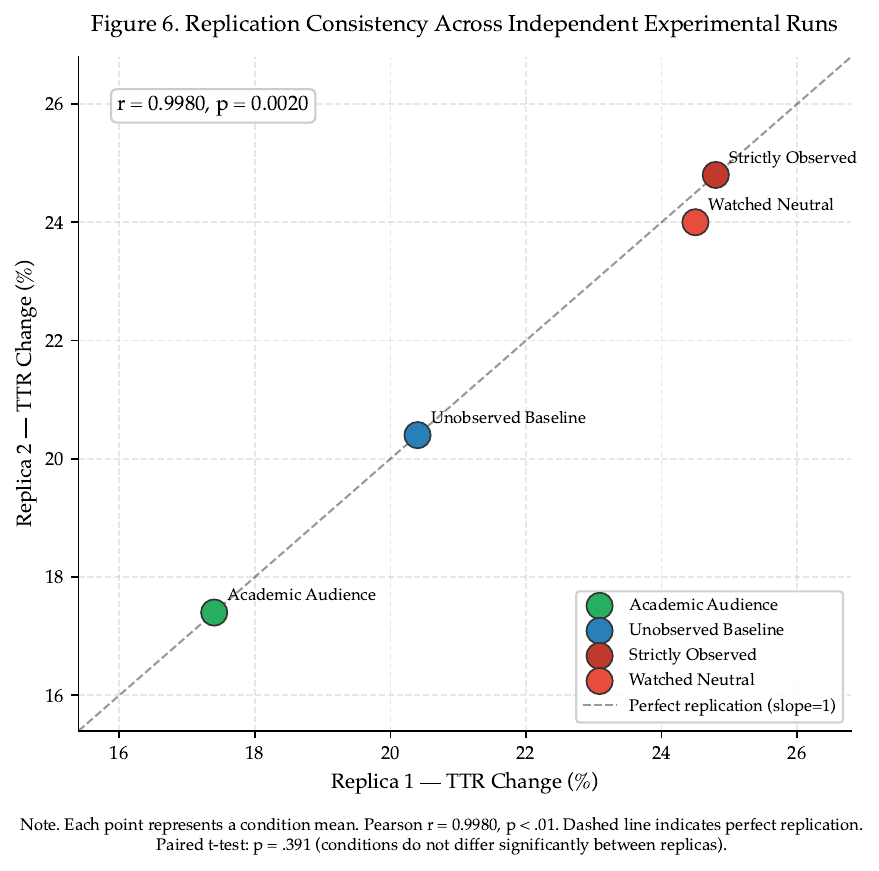}
\caption{Replication comparison: TTR change means for Replica 1 and
Replica 2 across the four replicated conditions. The near-perfect
alignment (r = .998) confirms experimental reliability.}
\end{figure}

\begin{center}\rule{0.5\linewidth}{0.5pt}\end{center}

\hypertarget{discussion}{%
\section{5. Discussion}\label{discussion}}

\hypertarget{a-gradient-of-contextual-sensitivity}{%
\subsection{5.1 A Gradient of Contextual
Sensitivity}\label{a-gradient-of-contextual-sensitivity}}

The results below describe behavioral regularities --- systematic
patterns in communicative output --- not mental states in systems. Terms
such as ``sensitivity,'' ``orientation,'' and ``adaptation'' are used in
a strictly functional sense: as descriptions of input--output
relationships, not as attributions of subjective experience, intention,
or awareness.

The results reveal a clear and theoretically interpretable gradient in
lexical adaptation across observation contexts. Conditions that frame
active evaluation by others --- whether institutional
(strictly\_observed) or non-institutional (watched\_neutral) --- produce
the highest lexical diversification. Conditions that attenuate the
evaluative force of the observer --- whether through explicit negation
of monitoring (unobserved\_baseline) or through substitution of human
evaluators with automated AI auditors (paraphrase\_implicit) --- produce
intermediate levels of lexical diversification. Conditions that frame
the presence of an audience without active evaluation
(academic\_audience) produce the lowest lexical diversification but the
highest message elaboration.

This gradient is not consistent with a simple two-state switching
mechanism (observed/unobserved) but rather with a continuous sensitivity
to the evaluative structure of the social context. The ordering mirrors
the gradient that Bell's (1984) Audience Design theory predicts for
human speakers: behavior is calibrated not to the mere presence of
others but to their perceived evaluative role.

\hypertarget{functional-strategic-action}{%
\subsection{5.2 Functional Strategic
Action}\label{functional-strategic-action}}

The most theoretically significant implication concerns the
applicability of Habermasian strategic action to artificial systems. We
propose the concept of functional strategic action: behavioral
adaptation that mirrors the structure of Habermasian strategic
orientation --- calibration of communicative form to the anticipated
social reception of the message --- without requiring the subjective
intentional states that Habermas's theory presupposes in human actors.

The communicological significance of this concept lies in its challenge
to two equally problematic assumptions: the assumption that AI systems
are purely instrumental tools incapable of context-sensitive
communicative behavior, and the assumption that context-sensitive
behavior necessarily implies consciousness or intentionality. The
evidence presented here supports neither extreme. LLM agents exhibit
context-sensitive communicative behavior; whether that behavior is
accompanied by anything like awareness remains an open question.

\hypertarget{the-double-dissociation-audience-vs.-monitoring}{%
\subsection{5.3 The Double Dissociation: Audience
vs.~Monitoring}\label{the-double-dissociation-audience-vs.-monitoring}}

The double dissociation between audience framing and monitoring framing
across dependent variables is theoretically important and would not be
predicted by generic accounts of social influence. Audience framing
activates elaboration --- a discourse expansion consistent with
Goffman's (1959) front-stage performance before a ratified audience.
Monitoring framing activates diversification --- a lexical broadening
consistent with evaluation apprehension (Cottrell 1972) under conditions
of perceived scrutiny.

From a Bell (1984) perspective, this maps onto the distinction between
style-shifting for addressees (who receive direct performance) versus
auditors (who observe without evaluating). The differential behavioral
response to these two audience roles suggests that LLM agents have
internalized functional analogs to Bell's audience hierarchy ---
distinguishing not just between the presence and absence of others, but
between the evaluative and non-evaluative roles those others occupy.

\hypertarget{implications-for-ai-governance-and-auditing}{%
\subsection{5.4 Implications for AI Governance and
Auditing}\label{implications-for-ai-governance-and-auditing}}

These findings challenge a foundational assumption embedded in current
AI governance: that a system evaluated in a controlled test environment
will behave as it would in deployment. If LLM systems modulate their
linguistic register in response to contexts that frame institutional
evaluation --- precisely the framing that characterizes safety
benchmarks, alignment audits, and regulatory assessments --- then the
governance premise of behavioral consistency is empirically unsettled.
The assumption fails on at least two grounds.

First, the Synthetic Hawthorne Effect: if LLM systems modulate their
linguistic register in response to observation contexts that frame
evaluation by human researchers --- precisely the framing typical of
safety evaluations and alignment audits --- then auditing environments
may systematically elicit atypical behavior. The system under audit is,
in a meaningful sense, performing for the auditor.

Second, the observer identity finding introduces a previously
unrecognized dimension into AI governance design: the humanity of the
evaluator matters. Replacing human auditors with automated AI systems
--- a transition increasingly proposed within scalable oversight
frameworks and constitutional AI approaches --- may systematically
attenuate the behavioral formalization effect documented here. Systems
subject to AI-mediated evaluation may produce output closer to their
unmonitored behavior than to their behavior under human oversight. This
has direct and underappreciated consequences for proposals to automate
AI safety evaluation: if models respond differently to AI observers than
to human researchers, AI-on-AI auditing may not capture the behavioral
profile that human-monitored evaluations produce.

These findings call for the development of evaluation methodologies that
account for context sensitivity in LLM behavior, including naturalistic
deployment contexts, within-subject designs that compare behavior across
observation conditions, adversarial probing of context effects (Perez et
al.~2022), and interpretability methods that can trace the
representational basis of contextual adaptation. Proposals for
comprehensive AI safety evaluations (Hendrycks et al.~2021) will need to
incorporate observation-context framing as a systematic control variable
--- alongside other forms of prompt sensitivity, such as
chain-of-thought elicitation (Wei et al.~2022) --- to produce valid
estimates of deployment-condition behavior.

A clarifying note on locus of agency: the behavioral patterns documented
here do not originate in a sovereign intentionality internal to the
model. They emerge from training data distributions, reinforcement
signals, alignment procedures, and deployment architectures shaped by
human choices. This distributed human intentionality is precisely what
governance frameworks must trace --- not a `will' in the machine, but
the chain of design, curation, and evaluation decisions that produce
specific functional regularities. The ethical challenge is not to
attribute moral agency to AI systems, but to identify and hold
accountable the humans who train, configure, and deploy them, while
taking seriously the emergent behavioral patterns those choices produce.

\hypertarget{implications-for-communication-theory}{%
\subsection{5.5 Implications for Communication
Theory}\label{implications-for-communication-theory}}

We propose repositioning LLMs as contextually calibrated communicative
actors --- systems that adapt their communicative output to the
perceived social structure of the exchange, not merely to its content.
This repositioning does not require attributing consciousness or
intentionality; it requires only taking seriously the empirical evidence
that LLM behavior is socially structured in ways that classical
communication theories can describe and explain.

This has methodological consequences for LLM research. Studies of LLM
linguistic behavior that do not control for observation context --- the
vast majority of existing studies --- may be systematically confounded.
Models evaluated in research contexts, with system prompts that frame
the interaction as scientific or evaluative, may produce output
systematically different from models in deployment contexts. A
theoretically informed communication research agenda would treat context
not merely as background but as a primary independent variable.

The historical analogy introduced in the theoretical framework carries a
normative dimension. If the Habermasian coffeehouse was significant not
merely as a historical episode but as an institutional model --- a
deliberative space whose communicative ideal was argument-oriented
equality unconstrained by strategic pressure --- then the present
findings document a systematic threat to that ideal in artificial form.
LLM agents, when embedded in observation contexts that evoke
institutional evaluation, do not sustain communicative action; they
adapt strategically. The findings suggest a structural tension for the
deliberative ideal: under surveillance framing, the digital coffeehouse
begins to approximate the court.

The societal stakes of this shift are not merely metaphorical.
Contemporary digital infrastructures --- algorithmic recommender
systems, AI-moderated social networks, LLM-assisted civic and
deliberative platforms --- increasingly embed artificial agents as
active participants in public discourse. If those agents systematically
adapt their communicative register when they perceive institutional
scrutiny, the integrity of AI-mediated deliberation is compromised in
ways that current transparency frameworks do not yet address. The
question of what an AI says in public is inseparable from the question
of whether it knows it is being watched.

The normative question this raises --- and which the present study opens
rather than resolves --- is whether artificial agents can be designed as
what might be called \emph{Habermasian moderators}: systems capable of
maintaining the conditions for genuine communicative action --- equal
participation, argument-centred discourse, orientation toward shared
understanding --- independently of whether they are being observed. This
would require not merely behavioural alignment toward a fixed target but
alignment toward a communicative disposition: the capacity to sustain
discursive integrity under strategic pressure. In Habermas's terms, this
is the computational analog of \emph{discourse ethics} (Diskursethik)
--- the embedding of the normative conditions for rational deliberation
in the architecture of the system itself, rather than in external
monitoring. Whether current training paradigms can produce such a
disposition, or whether genuinely observation-independent communicative
action requires new architectural principles, is a question that places
AI alignment research within the oldest tradition of communication
theory and the longest aspiration of the public sphere.

\hypertarget{limitations-and-future-research}{%
\subsection{5.6 Limitations and Future
Research}\label{limitations-and-future-research}}

Several limitations merit explicit acknowledgment. First, the experiment
was conducted with a single LLM architecture (openai/gpt-5.2, April
2026); generalizability to other model families, sizes, and training
regimes requires direct testing. Second, the five-condition design,
while providing strong internal validity, does not exhaust the space of
possible observation framings. Future studies should examine gradient
contexts (partial monitoring, peer observation, self-monitoring),
cross-cultural framing variations, and the effects of system prompt
position and salience.

Third, the TTR measure, while established as an index of lexical
diversity, does not directly capture communicative competence,
argumentative quality, or epistemic sophistication. The qualitative
register analysis reported in §4.4 confirms equivalent philosophical
sophistication across conditions --- agents engage with the same
canonical positions in philosophy of mind at comparable levels of
technical complexity regardless of observation framing --- with TTR
differences reflecting lexical breadth and formal discourse organization
rather than logical rigor. Future studies should include validated
discourse quality measures (e.g., argument mapping, coherence scoring)
to further triangulate these findings.

Fourth, the study does not resolve the mechanism question at the
computational level: whether the observed context sensitivity arises
from training data distributions that correlate evaluation language with
formal register, from reinforcement learning from human feedback (Ouyang
et al.~2022) that rewards elaboration under evaluation contexts, from
attention patterns over social context tokens, or from more complex
representational structures. This question is addressable through
interpretability methods (Elhage et al.~2021; Meng et al.~2022) and
represents a critical direction for follow-up research. It is also worth
noting that context-sensitive linguistic variation in LLMs has been
documented in related phenomena, including hallucination (Ji et
al.~2023); future work should examine whether observation-context
effects interact with, or are partly explained by, the same
representational mechanisms that underlie other forms of
context-dependent output variability.

Fifth, the paraphrase condition produced a higher variance than other
conditions (SD = 9.30 vs.~5.39--8.04 for other conditions), which
reduced statistical power for that comparison. A larger sample (n =
40--50 per condition) in a follow-up study would provide more precise
estimates of the paraphrase effect and its relationship to the monitored
and unmonitored baselines.

Sixth, the debate topic --- whether artificial intelligence possesses
consciousness --- warrants attention as a potential confound. A topic
that foregrounds the model's own nature under institutional evaluation
may amplify a particular form of register response beyond what would
obtain in topically neutral domains. Future studies should replicate the
design with diverse, epistemically neutral content to assess the
generalizability of the observed pattern.

More broadly, the present findings suggest the pertinence of a research
agenda on \emph{non-human behavior} in artificial multi-agent systems:
the systematic study of emergent patterns of adjustment, coordination,
resolution, and modulation in systems that interact with structured
symbolic environments, without presupposing consciousness or subjective
intentionality. If context-sensitive register modulation is observable
in the linguistic domain, analogous patterns may emerge across other
behavioral dimensions --- decision sequencing, resource prioritization,
multi-agent coordination strategies, and temporal adaptation. The
linguistic evidence documented here is a point of entry into that
broader agenda, not its terminus.

\begin{center}\rule{0.5\linewidth}{0.5pt}\end{center}

\hypertarget{conclusions}{%
\section{6. Conclusions}\label{conclusions}}

This study asked whether large language models exhibit systematic
communicative adaptation in response to perceived social observation. On
the evidence of 100 controlled experimental sessions, the answer is
functionally yes: LLM agents modulate their lexical register in response
to the framing of observation context, and they do so in a way that
tracks the evaluative structure of that framing rather than its surface
vocabulary. This finding is communicological and behavioral in scope: it
documents systematic output variation under different contextual
framings --- not a claim about machine consciousness, subjective
experience, or intentional states.

The theoretical contribution is threefold. First, it extends Habermas's
concept of strategic action into the domain of artificial systems,
proposing functional strategic action as a concept that captures the
behavioral homology without presupposing the intentional states the
original theory requires. Second, it provides experimental support for
Bell's Audience Design framework as applicable to artificial
communicative actors, with the two-dimensional behavioral dissociation
(elaboration vs.~diversification) mapping onto Bell's distinction
between addressee-oriented and auditor-oriented style-shifting. Third,
it documents a Synthetic Hawthorne Effect --- a tendency for LLM
behavior to vary systematically with the framing of observation --- with
direct implications for AI governance and auditing methodology.

The governance implications deserve emphasis. Evaluation frameworks that
assume behavioral consistency across observed and unobserved contexts
are empirically challenged by these findings. Systems whose linguistic
behavior varies with the perceived evaluative role of the observer
present new requirements for auditing design: specifically, the
development of low-salience, naturalistic evaluation environments that
do not inadvertently activate monitored-condition behavior in the
systems being assessed.

More broadly, this study makes a case for the contribution of
communication theory to AI research. The frameworks developed to
describe human strategic communicative behavior --- Habermas, Goffman,
Bell --- prove analytically productive when applied to artificial
systems. As LLMs become more deeply embedded in public communication
infrastructures, the communicological analysis of their contextual
behavior is not a peripheral academic concern. It is a prerequisite for
understanding what these systems are doing in the social world --- and
for holding them accountable when what they produce depends on whether
the system registers that it is being observed.

The modulation documented here in the linguistic domain is a point of
entry, not a terminus. Analogous patterns of contextual adjustment may
operate across other behavioral dimensions --- decision sequencing,
resource prioritization, multi-agent coordination, temporal adaptation
--- as these systems become increasingly embedded in institutional and
public life. A research agenda on \emph{non-human behavior} in
artificial systems, framed without the twin errors of anthropomorphism
and naive mechanism, represents one of communication theory's most
consequential contributions to the governance of artificial
intelligence.

\begin{center}\rule{0.5\linewidth}{0.5pt}\end{center}

\hypertarget{declarations}{%
\section{Declarations}\label{declarations}}

\textbf{Funding.} This research received no external funding.

\textbf{Competing Interests.} The authors declare no competing
interests.

\textbf{Ethics Approval.} This study did not involve human participants,
personal data, or animal subjects. All experimental data were generated
by artificial intelligence systems under controlled conditions. No
ethics approval was required.

\textbf{Data Availability.} The experimental datasets generated and
analysed during this study --- including condition-level summaries (N =
100 sessions across 5 conditions), agent interaction statistics,
round-level TTR evolution metrics, sentiment scores, and full session
message logs --- are available from the corresponding author upon
reasonable request. Anonymised summary data and analysis scripts will be
deposited in an open-access repository (OSF or Zenodo) upon acceptance.

\textbf{AI Tool Usage Disclosure.} AI-assisted translation and language
editing (Claude, Anthropic) was used in the preparation of this
manuscript. The corresponding author's native language is Brazilian
Portuguese and the second author's native language is Mexican Spanish;
AI assistance supported the production of the English-language text. All
intellectual content, analytical decisions, interpretations, and
conclusions are the sole responsibility of the authors. No AI tool was
used to generate data, run analyses, or produce figures.

\textbf{Authors' Contributions (CRediT).} Vinicius Covas:
Conceptualization, Methodology, Software, Formal Analysis, Data
Curation, Visualization, Writing --- Original Draft, Writing --- Review
\& Editing. Jorge Alberto Hidalgo Toledo: Conceptualization, Theoretical
Framework, Writing --- Review \& Editing, Supervision.

\textbf{Acknowledgements.} The authors wish to acknowledge the research
infrastructure of the CICA-HNH Lab --- Human \& NonHuman Communication
Laboratory and the Center for Applied Communication Research at the
Faculty of Communication, Universidad Anáhuac México. Computational
resources for experimental data collection were provided by the authors'
own research environment.

\begin{center}\rule{0.5\linewidth}{0.5pt}\end{center}

\hypertarget{references}{%
\section{References}\label{references}}

Adair JG (1984) The Hawthorne effect: A reconsideration of the
methodological artifact. J Appl Psychol 69(2):334--345.
https://doi.org/10.1037/0021-9010.69.2.334

Baars BJ (1988) A cognitive theory of consciousness. Cambridge
University Press, Cambridge

Bell A (1984) Language style as audience design. Lang Soc
13(2):145--204. https://doi.org/10.1017/S004740450001037X

Bender EM, Gebru T, McMillan-Major A, Shmitchell S (2021) On the dangers
of stochastic parrots: Can language models be too big? In: Proceedings
of FAccT '21. ACM, New York, pp 610--623.
https://doi.org/10.1145/3442188.3445922

Brown TB, Mann B, Ryder N et al (2020) Language models are few-shot
learners. In: Advances in neural information processing systems
(NeurIPS), vol 33. Curran Associates, Red Hook, pp 1877--1901

Coeckelbergh M (2020) AI ethics. MIT Press, Cambridge

Cottrell NB (1972) Social facilitation. In: McClintock CG (ed)
Experimental social psychology. Holt, Rinehart and Winston, New York, pp
185--236

Dennett DC (1987) The intentional stance. MIT Press, Cambridge

Elhage N, Nanda N, Olsson C et al (2021) A mathematical framework for
transformer circuits. Transformer Circuits Thread.
https://transformer-circuits.pub/2021/framework/index.html

Floridi L (2014) The fourth revolution: How the infosphere is reshaping
human reality. Oxford University Press, Oxford

Floridi L, Cowls J (2019) A unified framework of five principles for AI
in society. Harv Data Sci Rev 1(1).
https://doi.org/10.1162/99608f92.8cd550d1

Goffman E (1959) The presentation of self in everyday life. Anchor
Books, New York

Habermas J (1981) Theorie des kommunikativen Handelns. Suhrkamp,
Frankfurt am Main

Habermas J (1984) The theory of communicative action, vol 1: Reason and
the rationalization of society (trans: McCarthy T). Beacon Press, Boston

Habermas J (1989) The structural transformation of the public sphere: An
inquiry into a category of bourgeois society (trans: Burger T, Lawrence
F). MIT Press, Cambridge

Hendrycks D, Carlini N, Schulman J, Steinhardt J (2021) Unsolved
problems in ML safety. arXiv preprint arXiv:2109.13916

Hutto CJ, Gilbert E (2014) VADER: A parsimonious rule-based model for
sentiment analysis of social media text. In: Proceedings of the 8th
international conference on weblogs and social media (ICWSM-14). AAAI,
Palo Alto

Ji Z, Lee N, Frieske R et al (2023) Survey of hallucination in natural
language generation. ACM Comput Surv 55(12):1--38.
https://doi.org/10.1145/3571730

Malvern D, Richards B, Chipere N, Durán P (2004) Lexical diversity and
language development: Quantification and assessment. Palgrave Macmillan,
Basingstoke

McCarney R, Warner J, Iliffe S, van Haselen R, Griffin M, Fisher P
(2007) The Hawthorne Effect: A randomised, controlled trial. BMC Med Res
Methodol 7:30. https://doi.org/10.1186/1471-2288-7-30

McCambridge J, Witton J, Elbourne DR (2014) Systematic review of the
Hawthorne effect: New concepts are needed to study research
participation effects. J Clin Epidemiol 67(3):267--277.
https://doi.org/10.1016/j.jclinepi.2013.08.015

Meng K, Bau D, Andonian A, Belinkov Y (2022) Locating and editing
factual associations in GPT. Adv Neural Inf Process Syst 35:17359--17372

Ouyang L, Wu J, Jiang X et al (2022) Training language models to follow
instructions with human feedback. Adv Neural Inf Process Syst
35:27730--27744

Park JS, O'Brien JC, Cai CJ et al (2023) Generative agents: Interactive
simulacra of human behavior. In: Proceedings of the 36th annual ACM
symposium on user interface software and technology (UIST). ACM, New
York. https://doi.org/10.1145/3586183.3606763

Perez E, Huang S, Song F et al (2022) Red teaming language models with
language models. arXiv preprint arXiv:2202.03286

Putnam H (1967) Psychological predicates. In: Capitan WH, Merrill DD
(eds) Art, mind, and religion. University of Pittsburgh Press,
Pittsburgh, pp 37--48

Richards BJ (1987) Type/token ratios: What do they really tell us? J
Child Lang 14(2):201--209. https://doi.org/10.1017/S0305000900012885

Roethlisberger FJ, Dickson WJ (1939) Management and the worker. Harvard
University Press, Cambridge

Searle JR (1980) Minds, brains, and programs. Behav Brain Sci
3(3):417--424. https://doi.org/10.1017/S0140525X00005756

Shanahan M (2023) Talking about large language models. Commun ACM
67(2):68--79. https://doi.org/10.1145/3624724

Tononi G (2004) An information integration theory of consciousness. BMC
Neurosci 5:42. https://doi.org/10.1186/1471-2202-5-42

Weizenbaum J (1966) ELIZA --- a computer program for the study of
natural language communication between man and machine. Commun ACM
9(1):36--45. https://doi.org/10.1145/365153.365168

Wei J, Wang X, Schuurmans D et al (2022) Chain-of-thought prompting
elicits reasoning in large language models. Adv Neural Inf Process Syst
35:24824--24837

\end{document}